\documentclass[runningheads]{llncs}
\usepackage{amsmath}
\usepackage{amsfonts}
\usepackage{algorithm}
\usepackage[T1]{fontenc}
\usepackage{listings}
\usepackage{graphicx}
\usepackage{verbatim}
\usepackage{bm}
\usepackage{booktabs}
\usepackage{multirow}
\usepackage{subcaption}%
\usepackage{threeparttable}%
\usepackage[pagebackref,breaklinks,colorlinks]{hyperref}
\usepackage{url}
\usepackage{color}
\usepackage{xcolor}%
\usepackage{colortbl}%

\usepackage{svg}

\usepackage{tikz,siunitx}
\usetikzlibrary{shapes.geometric,shapes.symbols}
\newcommand{\tablestyle}[2]{\setlength{\tabcolsep}{#1}\renewcommand{\arraystretch}{#2}\centering\footnotesize}

\definecolor{tablered}{RGB}{205,51,51}
\definecolor{tablegreen}{HTML}{39b54a}
\definecolor{tableblue}{HTML}{4682B4}
\definecolor{tablegrey}{HTML}{808080}

\definecolor{figgrey}{RGB}{128,128,128}
\definecolor{figcyan}{RGB}{0,139,139}
\definecolor{figdarkred}{RGB}{64,0,0}
\definecolor{figred}{RGB}{205,51,51}
\definecolor{figgreen}{RGB}{113,198,113}

\definecolor{tomato}{HTML}{FF6347}
\definecolor{royalblue}{HTML}{4169E1}
\definecolor{springgreen}{HTML}{00FF7F}

\definecolor{simcolor1}{RGB}{214,96,77}
\definecolor{simcolor2}{RGB}{67,147,195}
\definecolor{rectcolor}{RGB}{27,120,55}

\definecolor{codegreen}{rgb}{0,0.6,0}
\definecolor{codegray}{rgb}{0.5,0.5,0.5}
\definecolor{codepurple}{rgb}{0.58,0,0.82}
\begin{document}
\title{Robust Noisy Pseudo-label Learning for Semi-supervised Medical Image Segmentation Using Diffusion Model}
\titlerunning{Robust Noisy Pseudo-label Learning Using Diffusion Model}
%
\begin{comment}  %% Removed for anonymized MICCAI 2025 submission
\author{First Author\inst{1}\orcidID{0000-1111-2222-3333} \and
Second Author\inst{2,3}\orcidID{1111-2222-3333-4444} \and
Third Author\inst{3}\orcidID{2222--3333-4444-5555}}
%
\authorrunning{F. Author et al.}
\institute{Princeton University, Princeton NJ 08544, USA \and
Springer Heidelberg, Tiergartenstr. 17, 69121 Heidelberg, Germany
\email{lncs@springer.com}\\
\url{http://www.springer.com/gp/computer-science/lncs} \and
ABC Institute, Rupert-Karls-University Heidelberg, Heidelberg, Germany\\
\email{\{abc,lncs\}@uni-heidelberg.de}}

\end{comment}

%\author{Anonymized Authors}  %% Added for anonymized MICCAI 2025 submission
%\authorrunning{Anonymized Author et al.}
%\institute{Anonymized Affiliations \\
%    \email{email@anonymized.com}}
\author{Lin Xi\inst{1}\orcidID{0000-0001-6075-5614} \and Yingliang Ma\thanks{Corresponding author: Y. Ma}\inst{1,2}\orcidID{0000-0001-5770-5843
	} \and Cheng Wang\inst{1} \and Sandra Howell\inst{2} \and Aldo Rinaldi\inst{2} \and Kawal S. Rhode\inst{2}}
\authorrunning{L. Xi et al.}
\institute{University of East Anglia, Norwich NR4 7TJ, United Kingdom \and King's College London, London SE1 7EH, United Kingdom\\\email{\{l.xi,yingliang.ma\}@uea.ac.uk}}

\maketitle              % typeset the header of the contribution
\begin{abstract}
Obtaining pixel-level annotations in the medical domain is both expensive and time-consuming, often requiring close collaboration between clinical experts and developers. Semi-supervised medical image segmentation aims to leverage limited annotated data alongside abundant unlabeled data to achieve accurate segmentation. However, existing semi-supervised methods often struggle to structure semantic distributions in the latent space due to noise introduced by pseudo-labels. In this paper, we propose a novel diffusion-based framework for semi-supervised medical image segmentation. Our method introduces a constraint into the latent structure of semantic labels during the denoising diffusion process by enforcing prototype-based contrastive consistency. Rather than explicitly delineating semantic boundaries, the model leverages class prototypes centralized semantic representations in the latent space as anchors. This strategy improves the robustness of dense predictions, particularly in the presence of noisy pseudo-labels. We also introduce a new publicly available benchmark: Multi-Object Segmentation in X-ray Angiography Videos (MOSXAV), which provides detailed, manually annotated segmentation ground truth for multiple anatomical structures in X-ray angiography videos. Extensive experiments on the EndoScapes2023 and MOSXAV datasets demonstrate that our method outperforms state-of-the-art medical image segmentation approaches under the semi-supervised learning setting. This work presents a robust and data-efficient diffusion model that offers enhanced flexibility and strong potential for a wide range of clinical applications.

% Authors must provide keywords and are not allowed to remove this Keyword section.

\end{abstract}
\section{Introduction}\label{sec:intro}

Medical image segmentation, which involves pixel-wise classification, is a critical dense prediction task that plays a vital role in enhancing the accuracy of disease diagnosis, monitoring, and assessment. In recent years, learning-based approaches \cite{UNet,AttnUNet,SwinUNet,MedSAM,Ma2022TBME,Ma2022Edge,Ma2024BIS} have significantly outperformed traditional methods, with fully supervised image segmentation achieving higher accuracy due to the abundance of annotated data. In other words, large-scale image datasets \cite{BRATS,AbdomenCT1K,Cholecseg8k,Ma2018MP,Ma2021PMB} with manual annotations are typically required to train robust and generalizable deep neural networks for dense prediction tasks, \emph{e.g.}, image segmentation. However, obtaining pixel-wise annotations is challenging, as it both time-consuming and labor-intensive. Such a massive annotation cost has motivated the community to develop semi-supervised learning methods \cite{DiffRect,PAMT,SSMIS}.

Given that unlabeled data is typically abundant in practice, semi-supervised learning has emerged as a compelling approach for medical image segmentation. It leverages limited labeled data in conjunction with large amounts of pseudo-labeled data to iteratively train the segmentation model \cite{CPCSAM,DiffRect}. In this paradigm, pseudo labels are generated for unlabeled images using a reliable model, and these pseudo-labeled samples are effectively incorporated into training by minimizing their prediction entropy. Nevertheless, \textbf{semantic misalignment} remains a significant challenge in semi-supervised learning. Most existing methods \cite{CPS,FixMatch} rely primarily on consistency regularization and auxiliary supervision at the output mask level, implicitly enforcing semantic consistency under perturbations (\emph{e.g.}, pseudo labels). However, these approaches often result in overlapping semantic representations in the latent space and overlook distinct semantic boundaries, thereby limiting generalization.

To address the above issue, some previous works have proposed modeling data distribution with deep generative models, including Variational Autoencoders (VAEs) \cite{VAE}, Generative Adversarial Networks (GANs) \cite{GAN}, and specialized medical image segmentation diffusion models \cite{MedSegDiff,DiffRect}. Among these generalist approaches, diffusion models \cite{DiffRect} have demonstrated significant potential in alleviating this problem by formulating complex data distributions probabilistically. A prominent branch treats dense prediction tasks as a label-denoising problem \cite{DDP,MaskDiff}, employing variational inference to progressively generate predictions from noisy data. While diffusion models are proven capable of capturing the underlying distribution of each semantic category, their full potential to distinctly shape the latent structure of semantic labels remains to be discovered. The precise delineation of semantic boundaries and distinct domains within each semantic representation is crucial for improving overall performance in semi-supervised dense prediction tasks by eliminating ambiguity. These observations motivate our investigation into whether diffusion-based deep generative models can accurately capture and align the precise distribution of semantic labels, enabling the progressive rectification of pseudo-labels for enhanced supervision.

In this paper, we propose a novel diffusion model for semi-supervised medical image segmentation. Specifically, we first use embedding layers to encode dense labels and map these features into a semantic latent space via project layers. To structure the latent space, we introduce explicit class prototypes as fixed anchors in the embedding space and directly optimize the feature representations using a contrastive loss. These non-learnable prototypes guide the model beyond simply optimizing for prediction accuracy. The denoised embedding features are then processed by a diffusion decoder, guided by visual conditions. 

Our main contributions are as follows: 1) We propose a novel diffusion-based framework for semi-supervised medical image segmentation, which performs dense label denoising through a diffusion process by reformulating dense prediction as a label denoising task. 2) We introduce a prototype-anchored contrastive learning to structure the latent space of semantic labels, enhancing the quality and robustness of segmentation predictions. 3) We present a new benchmark dataset, Multi-Object Segmentation in X-ray Angiography Videos (MOSXAV) \footnote{https://github.com/xilin-x/MOSXAV}, which focuses on the segmentation of multiple objects in X-ray angiography videos and provides high-quality, manually annotated ground truth labels.

To validate these contributions, the proposed method is evaluated on one public benchmark and a newly collected benchmark dataset. Experimental results demonstrate that our algorithm outperforms state-of-the-art (SoTA) medical image segmentation methods under the semi-supervised learning setting.

\section{Method}\label{sec:met}

\subsection{Preliminaries}\label{subsec:prelim}

%\begin{equation}\label{eq:forward}
%	z_{t}=\sqrt{\gamma(t)}z_{0}+\sqrt{1-\gamma(t)}\epsilon,
%\end{equation}

Diffusion models \cite{pmlr_sohl,DDPM,DDIM}, which consist of a forward noising process and a reverse denoising process, are widely used in generative tasks. The forward noising process gradually adds noise to the data sample to generate a noisy sample $z_{t}$, which can be formulated as $z_{t}$=$\sqrt{\gamma(t)}z_{0}$+$\sqrt{1-\gamma(t)}\epsilon$, where $\epsilon$ is the Gaussian noise and $t$$\in$$\{0,1,...,T\}$ indicates the time steps. $\gamma(t)$ is a monotonically decreasing function to control the signal-to-noise ratio and the degree of corrosion. In the forward process, the original data $z_{0}$ is iteratively broken towards the pure Gaussian noise $z_{T}$. At the training stage, a denosing network $f_{\theta}(z,t)$ parameterized by $\theta$ is trained to predict $z_{0}$ from $z_{t}$ by minimizing an objective function, which is a $l_{2}$ loss most of the time. At the inference stage, the diffusion models perform the reverse denoising process. The neural network follows a Markovian way that recovers $z_{0}$ from the pure Gaussian noise $z_{T}$ iteratively. Specifically, the process of $z_{T}$$\rightarrow$$z_{T-\delta}$$\rightarrow$$...$$\rightarrow z_{0}$ is achieved by applying the denoising network to $z_{0}$ and then using the predicted $\tilde{z}_{0}$ to make the transition to $z_{t-\delta}$ iteratively.

In perception tasks, the diffusion models usually take the feature $\mathbf{x}$ as conditions to perform denoising. For example, in the semantic segmentation task, the diffusion models take both the noisy segmentation label $z_{t}$ and conditional feature $\mathbf{x}$ as input to perform the denoising, which can be formulated as follows:
\begin{equation}\label{eq:condiff}
	q_{\theta}(z_{0:T}|\mathbf{x})=q(z_{T})\prod_{t=0}^{T}q_{\theta}(z_{t-1}|z_{t},\mathbf{x}),
\end{equation}
where $q_{\theta}(\cdot)$ is implemented by the transition rule based on denoising network $f_{\theta}(z,t,\mathbf{x})$ that takes $\mathbf{x}$ as conditional input. Our method is based on the conditional diffusion models to perform the medical image segmentation. We propose a label encoding that generates pixel-wise dense label embedded maps as input to the diffusion decoder. We utilize the visual features as the condition.

\subsection{Architecture}

\begin{figure*}[!tb]
	\begin{center}
		\includegraphics[width=1.0\linewidth]{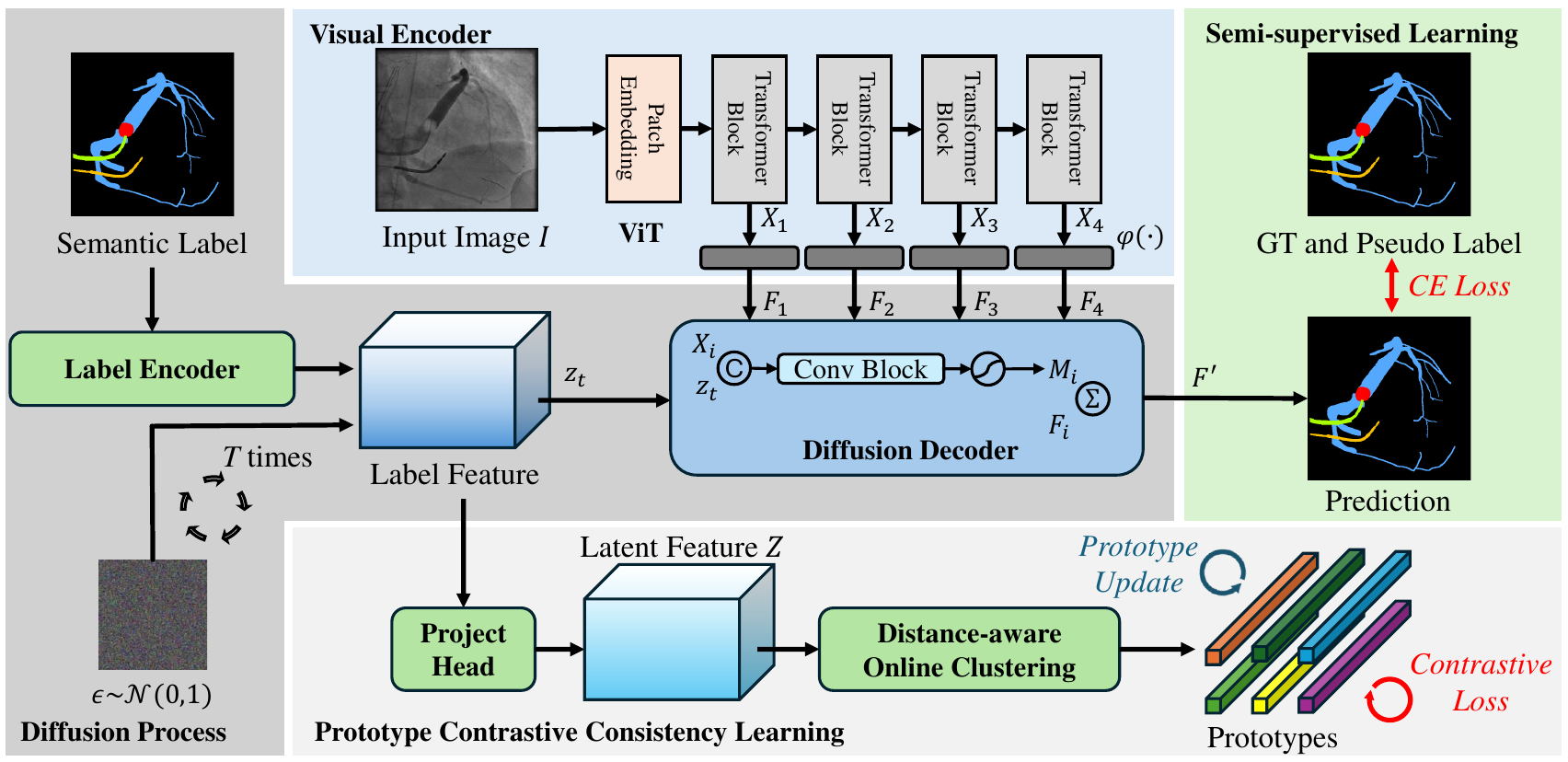}
	\end{center}
	\captionsetup{font=small}
	\caption{Overview of the proposed diffusion-based semi-supervised segmentation model.}
	\label{fig:framework}
\end{figure*}

The overall framework of our model is presented in Fig. \ref{fig:framework}. which comprises a visual encoder, a label encoder, a latent projector, and a diffusion decoder.

\noindent\textbf{Visual encoder}. The visual encoder takes an image $\bm{I}$ as input and extracts multi-level features $\bm{F}_{i}$$\in$$\mathbb{R}^{C\times H\times W}$ ($i$$\in$$\{1,2,...,N\}$) for diffusion decoder as visual condition to guide denoising. $H$, $W$, and $C$ denote the height, width, and channels of one-level feature, respectively. $N$ indicates the total number of levels of visual encoder. Specifically, we first utilize a backbone ViT \cite{ViT} for visual branch to extract visual features. We select features from different layers of the backbone, denoted as $\bm{X}_{l}$ ($l$$\in$$\{l_{1},l_{2},...,l_{N}\}$), where $l$ is the layer index. The $N$-level features are fed into visual-specific branches to generate the visual-specific multi-level features $\bm{F}_{i}$. The visual-specific branch $\varphi(\cdot)$ contains two stacked convolutional blocks, each of which consists of a 3$\times$3 convolution followed by a batch normalization layer and a GeLU activation layer and a 1$\times$1 convolution. Moreover, to learn discriminative multi-level visual-specific features, we also design visual-specific auxiliary heads to generate intermediate predictions. These intermediate predictions are supervised by the corresponding semantic ground-truth and pseudo labels $M_{j}$ ($j$$\in$$\{\textrm{gt}, \textrm{pseudo}\}$) and the visual-specific branches will be updated by the gradient from segmentation prediction. The output of each visual-specific branch acts as the conditions for the diffusion decoder.

\noindent\textbf{Label and latent encoder.} We first encode the discrete, dense semantic label map into a latent feature space using a label encoder to capture semantic representations. Given the discrete nature of semantic segmentation labels, we begin by converting them into one-hot representations. The label encoder consists of a 1$\times$1 convolutional layer and a 3$\times$3 convolutional layer that transforms the dense labels into label-specific encoded features $z$. This output $z$ is then normalized to the range [$-1,+1$], and a scaling factor $s$ is applied to control the signal-to-noise ratio. This scaling increases the difficulty of the denoising task, thereby encouraging the diffusion decoder to learn more informative representations. To simulate the forward diffusion process, Gaussian noise is added to the label-specific encoded features $z$, producing the corrupted representation $z_{t}$. In the forward noising process, the corruption intensity is controlled by $\gamma(t)$$\in$[$0,1$], which decreases over time $t$. Following previous works \cite{DDP,ImprovedDDPM}, we adopt a cosine schedule for $\gamma(t)$ to modulate the noise level throughout the diffusion process.

Subsequently, we apply a latent encoder composed of a lightweight convolutional layer that performs a non-linear projection, denoted as $\phi:\mathbb{R}^{H\times W \times C}\mapsto\mathbb{R}^{H\times W\times D}$, producing the latent embedding $\bm{Z}$=$\phi(z)$. This projection step is commonly used in contrastive learning over convolutional feature maps \cite{ChenK0H20} to obtain more discriminative representations.

\noindent\textbf{Diffusion decoder} The decoder takes the noisy label-specific encoded features $z_{t}$ as input and the visual-specific multi-level features $\bm{F}_{i}$ as conditions. We also use the features $\bm{X}_{l}$ from the backbone different layers to help model the semantic space. The noisy label map $z_{t}$ is concatenated with the features $\bm{X}_{l}$ and sent to the diffusion decoder to perform level interaction successively with $\bm{F}_{i}$. Specifically, in the level-interaction phase, we perform the interaction at each level. We utilize 2 convolutional blocks (Conv-BatchNorm-GeLU-Conv) and a convolutional layer to map the concatenated feature map $z_{t}$ and $\bm{X}_{l}$ to each feature level and the channel number is transformed from $C$ to $1$. The multi-level mask maps are generated by applying a Sigmoid function to the output features of the convolutional block at each level, denoted as $\bm{M}_{i}$$\in$$\mathbb{R}^{1\times H \times W}$ ($i$$\in$$\{1,2,...,N\}$). Then, we use $\bm{M}_{i}$ to generate the final aggregated features by $\bm{F}'$=$\sum_{i=1}^{N}\bm{M}_{i}\cdot\bm{F}_{i}$. The aggregated feature $\bm{F}'$ is sent to the prediction branch to generate the final prediction, where the prediction branch contains three convolutional blocks. The predictions are then encoded as mentioned in $\S$ \ref{subsec:prelim} to generate the predicted $\tilde{z}_{0}$.

\subsection{Prototype Contrastive Consistency Learning}

Prototype-based learning represents semantic classes using class-specific prototypes and performs classification by comparing inputs to these representative prototypes. Let $\bm{p}_{c}$ ($c$$\in$${1,2,...,C}$) denote a set of prototypes corresponding to semantic class $c$. Rather than adopting a parametric prototype learning approach for constraint in semantic latent space, we introduce a non-parametric prototype contrastive learning framework. Specifically, we employ a set of non-learnable class prototypes and apply a contrastive loss directly to these prototypes via a prototype-anchored metric learning strategy. This method enables the latent semantic space to be structured more effectively by incorporating known inductive biases, such as \textbf{intra-class compactness} and \textbf{inter-class separability}, which are often overlooked in traditional approaches. The label-based contrastive loss encourages features of pixels from the same class to cluster closely, while enforcing separation between features from different classes. During training, since the prototypes are representative of the dataset as a whole, these inductive biases can be directly imposed as optimization objectives. This enables us to explicitly shape the embedding space beyond merely optimizing for prediction accuracy.

\noindent\textbf{Inter-class compactness.} We begin by grouping the latent embeddings $\bm{Z}$ into $K$ prototypes $\{\bm{p}_{c,k}\}_{k=1}^{K}$ for each semantic class $c$ in an online clustering manner. After processing all samples in the current batch, each pixel embedding $\bm{i}$$\in$$\bm{Z}$ is assigned to the $k$-th prototype of class $c$, where the assignment is determined by $k$=$\mathrm{arg}\max_{k}\{l\}_{k=1}^{k}$ with $l$=$\left\langle\bm{i},\bm{p}_{c,k}\right\rangle$ and $\left\langle\cdot,\cdot\right\rangle$ denotes a similarity metric, \emph{e.g.}, the cosine similarity. This prototype assignment leads to a training objective aimed at maximizing the posterior probability of the correct prototype assignment. This can be viewed as a inter-class prototype contrastive learning loss:
\begin{equation}\label{eq:inter}
	\mathcal{L}_{\mathrm{inter}}=-\log\frac{\exp(\bm{i}^\top\bm{p}_{c,k}/\tau)}{\exp(\bm{i}^\top\bm{p}_{c,k}/\tau)+\sum_{\bm{p}^{-}\in\mathcal{P}^{-}}\exp(\bm{i}^\top\bm{p}^{-}/\tau)},
\end{equation}
where $\mathcal{P}^{-}$ denotes the set of prototypes excluding those belonging to the target class $c$, and the temperature parameter $\tau$ regulates the concentration of the similarity distribution.

\noindent\textbf{Intra-class separability.} To further reduce intra-class variation, \emph{i.e.}, to encourage pixel features assigned to the same prototype to form compact clusters, we introduce a loss that regularizes the learned representations by directly minimizing the distance between each embedded pixel and its assigned prototype:
\begin{equation}\label{eq:intra}
	\mathcal{L}_{\mathrm{intra}}=(1-\bm{i}^\top\bm{p}_{c,k})^{2}.
\end{equation}
Notably, both $\bm{i}$ and $\bm{p}_{c,k}$ are $\ell_{2}$-normalized. This normalization ensures that the training objective focuses on angular similarity, minimizing intra-class variation while preserving separation between features assigned to different prototypes.

\subsection{Training and Inference}

During training, we first generate label-specific encoded features $z$, to which noise is added to obtain the noisy representation $z_t$. This noisy feature is then fed into the diffusion decoder, and the model is trained to perform the denoising. The overall procedure is outlined in Algorithm \ref{alg:training}. During inference, the diffusion model starts from an initial Gaussian noise and iteratively denoises the input to approximate the ground truth. This inference process is detailed in Algorithm~\ref{alg:inference}.

\begin{figure*}[!tb]
	\begin{minipage}{.49\linewidth}
		\begin{algorithm}[H]
			\caption{Training stage}
			\label{alg:training}
\begin{lstlisting}[language=Python]
def train(images, gts, p):
    """
    images: [B, 3, H, W]
    gts:    [B, 1, H, W]
    p:      [(CLS + 1) * SUB_CLS, D]
    """
    # Encode multi-level visual-specific features
    x = visual_encoder(images)
    f = varphi(x)  # visual-specific branches
    # Encoding gts
    z = label_encoder(gts)
    z_latent = latent_encoder(z)  # D-dimension

    # Compute prototype and assignment
    p, l = online_cluster(z_latent, gts, p)

    # Corrupt the label feature according to t
    t = randint(0, T)  # timestep
    eps = normal(mean=0, std=1) # noise as N(0,1)
    z_crpt = sqrt(alpha_cumprod(t)) * z + sqrt(1 - alpha_cumprod(t)) * eps

    # Predict and compute loss
    preds = diffusion_decoder(z_crpt, x, f, t)
    # Using CE, inter and intra loss to train
    loss = ce_loss(gts, preds) + 1e-2 * inter_loss(z_latent, p, l) + 1e-3 * intra_loss(z_latent, p, l)

    return loss
\end{lstlisting}
		\end{algorithm}
	\end{minipage}
	\begin{minipage}{.49\linewidth}
		\begin{algorithm}[H]
			\caption{Inference stage}
			\label{alg:inference}
\begin{lstlisting}[language=Python]
def infer(images, steps):
    """
    images: [B, 3, H, W]
    steps:  number of sample steps
    """
    # Extract multi-level visual-specific features
    x = visual_encoder(images)
    # Visual-specific branches
    f = varphi(x)

    # Generate noisy map
    m_t = normal(mean=0, std=1)

    for step in range(steps):
        # Generate time interval
        t_now = 1 - step / steps
        # Asymmetirc time sampling
        t_next = max(1 - (1 + step + t_diff) / steps, 0)

        # Predict pred from m_t
        preds = diffusion_decoder(m_t, x, f, t_now)

        # Encoding preds with label encoder
        z = label_encoder(preds)

        # Estimate map_t at t_next
        m_t = ddim(m_t, z, t_now, t_next)

    return preds
\end{lstlisting}
		\end{algorithm}
	\end{minipage}
\end{figure*}

We adopt DDIM \cite{DDIM} as the update rule for the noisy map. After predicting $\tilde{z}_{0}$ at each step, we apply the reparameterization trick to generate the noisy label-specific features for the subsequent step. Following the approaches in \cite{DDP}, we employ asymmetric time intervals during inference. These time intervals are controlled by \texttt{t\_diff} in Algorithm \ref{alg:inference}, which is empirically set to 1.

\section{Experiments}\label{sec:exp}

\subsection{Experimental Setup}

\noindent\textbf{Datasets.} We evaluate our model using two datasets. First, we employ the public Endoscapes2023 dataset \cite{Endoscapes2023}, which includes three sub-tasks: surgical scene segmentation, object detection, and critical view of safety assessment. For our experiments, we use the segmentation subset, which comprises 493 annotated frames extracted from 50 laparoscopic cholecystectomy videos. The dataset is officially split into training, validation, and test sets in a 3:1:1 ratio, respectively.

Additionally, we introduce a new benchmark dataset named MOSXAV, comprising 62 X-ray angiography video sequences: 40 from the CADICA dataset \cite{CADICA} and 22 collected from cardiac resynchronization therapy procedures performed at two hospitals. These videos capture the injection and flow of contrast agents through the coronary arteries or coronary sinus along the heart surface. Each video contains 33–70 frames at a resolution of 512$\times$512. Vascular regions were annotated by experienced radiologists, with one or two key frames labeled per video—specifically when the contrast agent is most visible. For training and validation, dense annotations are provided every 5 frames across 50 sequences (totaling 2,335 frames). The test set includes 488 frames, all of which are fully annotated.

\noindent\textbf{Evaluation metric.} Following conventions \cite{UNet,AttnUNet}, mean intersection-over-union (mIoU) is adopted for evaluation.

\noindent\textbf{Implementation Details.} We utilize ViT-Large \cite{ViT} as the visual backbone for our main experiments and ViT-Base for all ablation studies, both producing features with a stride of 16. All models are trained for 40,000 iterations with a batch size of 8. The network is optimized using the Adam optimizer ($\beta_{1}$=0.9, $\beta_{2}$=0.999), with an initial learning rate of 5e-4 and a weight decay of 1e-6 for both the Endoscapes2023 and MOSXAV datasets and a polynomial learning rate scheduler is employed. The semantic segmentation task is supervised using the cross-entropy loss. To generate pseudo masks for Endoscapes2023, we first use object detection bounding boxes as prompts for the MedSAM \cite{MedSAM}. The initial masks produced by MedSAM are then refined using SAMRefiner \cite{SAMRefiner} to obtain higher-quality pseudo labels. All the experiments are trained with 2 NVIDIA RTX 6000 GPUs for 40,000 iterations.

\subsection{Main Results}

To comprehensively evaluate the superiority of our proposed method, we show a comparison of SoTA results on the Endoscapes2023, \textit{val} and \textit{test} sets of MOSXAV in Table \ref{tab:SoTA}. Our method is evaluated alongside several publicly available baseline segmentation methods from the computer vision community. Note that the ground truth is obtained from the \textit{Seg50} subset, while the pseudo labels are generated for the \textit{BBox201} subset of the Endoscapes2023 dataset.

\begin{table*}[!tb]
	\centering
	%	\tiny
	\captionsetup{font=small}
	\caption{Quantitative results on the Endoscapes2023 and MOSXAV datasets, with the best performance scores highlighted in \textbf{bold}. The rows labeled GT, Pseudo, and GT + Pseudo correspond to models trained using ground truth labels, pseudo labels, and a combination of both, respectively.}
	\label{tab:SoTA}
	\begin{threeparttable}
		\resizebox{1.0\linewidth}{!}{%
						\tablestyle{5pt}{1.02}
			\begin{tabular}{@{\hskip 4pt}r|c|c|c|c|c@{\hskip 4pt}}\toprule
				\multirow{4}{*}{Methods~~~~~~} & \multicolumn{3}{c|}{Endoscapes2023} & \multicolumn{2}{c}{MOSXAV} \\
				& \multicolumn{3}{c|}{\textit{val}}        & \textit{val}          & \textit{test}        \\\cline{2-6} 
				& GT    & Pseudo  & GT + Pseudo  & \multicolumn{2}{c}{GT}     \\\cline{2-6} 
				& mIoU $\uparrow$ (\%)  & mIoU $\uparrow$ (\%)    & mIoU $\uparrow$ (\%)         & mIoU $\uparrow$ (\%)         & mIoU $\uparrow$ (\%)        \\\midrule
				U-Net \cite{UNet} &   41.44   &    44.09      &    45.64          &     67.58         &  27.81 \\ 
				Attention U-Net \cite{AttnUNet} &  42.13    &    45.06      &     45.47         &      67.91        & 32.08  \\
				TransU-Net \cite{TransUNet} &   45.00   &     45.06     &  46.05            &      68.22        & 30.91  \\
				CMU-Net \cite{CMUNet} &  45.63    &     46.23     &     46.92         &       68.36       & 27.95  \\
				CMU-NeXt \cite{CMUNeXt} &  39.95     &  40.97        &   40.61           &     64.11         & 24.85  \\
				SwinU-Net \cite{SwinUNet} &   36.65   &    37.23      &    37.83          &     63.46         &  22.67 \\
				MedSegDiff \cite{MedSegDiff} &  50.72    &  51.18        &    51.53          &     69.59         &  39.93	 \\\midrule
				Ours &   \textbf{55.33}   &     \textbf{56.04}     &        \textbf{56.68}      &        \textbf{73.48}      &  \textbf{43.04} \\\bottomrule        
			\end{tabular}
		}
	\end{threeparttable}
\end{table*}%

\noindent\textbf{Endoscapes2023 dataset.} Our method significantly outperforms baseline semantic segmentation approaches under the semi-supervised learning setting (\emph{i.e.}, GT + Pseudo), notably surpassing the diffusion-based model MedSegDiff \cite{MedSegDiff} by a margin of +5.15\%. Furthermore, our method also achieves the best performance when trained solely on ground truth or pseudo labels. 

\noindent\textbf{MOSXAV \textit{val} and \textit{test} sets.} Table \ref{tab:SoTA} shows that our method outperforms all baseline semantic segmentation methods on the \textit{val} set. More importantly, on the challenging \textit{test} set, it consistently surpasses MedSegDiff, achieving an mIoU improvement from 39.93\% to 43.04\%.

\subsection{Ablation Study}

To demonstrate the effectiveness of the different components in our model, we perform an ablation study on the Endoscapes2023 dataset.

\begin{table*}[!tb]
	\centering
	\captionsetup{font=small}
	\caption{Ablation studies on the Endoscapes2023 dataset.}
	\label{tab:abla}
	\resizebox{0.95\textwidth}{!}{
		\subfloat[{Training Objective $\mathcal{L}$} \label{table:abla:loss}]{
			\tablestyle{8pt}{1.05}
			\begin{tabular}{@{\hskip 4pt}ccc|c@{\hskip 4pt}}
				\toprule
				$\mathcal{L}_{\mathrm{CE}}$ & $\mathcal{L}_{\mathrm{inter}}$ & $\mathcal{L}_{\mathrm{intra}}$ & mIoU $\uparrow$ (\%) \\\midrule
				$\checkmark$ &        &        &    53.04  \\
				$\checkmark$ &   $\checkmark$     &        &    54.22  \\
				$\checkmark$ &        &   $\checkmark$     &   54.04   \\
				$\checkmark$ &    $\checkmark$    &     $\checkmark$   & \textbf{54.47} \\\bottomrule
			\end{tabular}%
		}\hfill\hspace{10pt}
		\subfloat[{Prototype Number $K$}\label{table:abla:protonum}]{%
			\tablestyle{8pt}{1.05}
				\begin{tabular}{@{\hskip 4pt}c|c@{\hskip 4pt}}
				\toprule
				\# Prototype & mIoU $\uparrow$ (\%) \\\midrule
				$K$=1  &    54.29  \\
				$K$=5  &    54.37  \\
				~$K$=10 &   \textbf{54.47}   \\
				~$K$=20 & 54.46\\\bottomrule
			\end{tabular}%
		}\hfill
	}
\end{table*}

\noindent\textbf{Effectiveness of Contrastive Loss.} To verify the effectiveness of our overall training objective, we progressively incorporate contrastive learning components defined in Eqs. \ref{eq:inter} and \ref{eq:intra} into the final loss function. As shown in Table \ref{table:abla:loss}, the model with $\mathcal{L}_{\mathrm{CE}}$ alone achieves an mIoU score of 53.04\%. Combing all the losses together leads to the best performance, yielding an mIoU score of 54.47\%.

\noindent\textbf{Prototype Number Per Class $K$.} Table \ref{table:abla:protonum} presents the performance of our approach concerning the number of prototype per class. For $K$=1, we directly represent each class as the mean embedding of its pixel samples. As a result, we set $K$=10 for a better trade-off between accuracy and computation cost.

\section{Conclusion}\label{sec:con}

In this paper, we propose a novel diffusion-based framework for semi-supervised medical image segmentation that enhances the robustness of dense predictions, particularly in the presence of noisy pseudo-labels. Our approach introduces a prototype contrastive consistency constraint into the latent structure of semantic labels during the denoising diffusion process. Instead of explicitly delineating semantic boundaries, the model utilizes class prototypes—centralized semantic representations in the latent space—as anchors to guide learning. Additionally, we introduce MOSXAV, a new publicly available benchmark dataset that provides detailed, manually annotated ground truth for multiple anatomical structures in X-ray angiography videos. Extensive experiments demonstrate that our method consistently outperforms SoTA medical image segmentation approaches under the semi-supervised learning setting.

\begin{credits}
	\subsubsection{\ackname} This work was supported by EPSRC UK (EP/X023826/1) and MRC impact fund (University of East Anglia).
	\subsubsection{\discintname} The authors have no competing interests to declare that are relevant to the content of this article.
\end{credits}

\bibliographystyle{splncs04}
\bibliography{mybib}
\end{document}